\documentclass[10pt, letter,twocolumn]{IEEEtran}

\usepackage{algorithm,algpseudocode}
\usepackage[dvips]{color}
\usepackage{comment}
\usepackage{epsf}
\usepackage{times}
\usepackage{epsfig}
\usepackage{float}
\usepackage{graphicx}
\usepackage{epstopdf}
\usepackage{algorithm}
\usepackage{amsmath}
\usepackage{amssymb}
\usepackage{amsxtra}
\usepackage{multirow}
\usepackage{mathtools}
\usepackage{cite}	
\usepackage{color}
\usepackage{bbm}
\usepackage{subcaption}
\usepackage[font={small}]{caption}
\title{An Improved Dilated Convolutional Network for Herd Counting in Crowded Scenes} 
\author{
\IEEEauthorblockN{\large Soufien Hamrouni$^1$, Hakim Ghazzai$^1$, Hamid Menouar$^2$, and Yehia Massoud$^1$\\
\IEEEauthorblockA{$^1$School of Systems \& Enterprises, Stevens Institute of Technology, Hoboken, NJ, USA}\\
\IEEEauthorblockA{$^2$Qatar Mobility Innovations Center (QMIC), Qatar University, Doha, Qatar}
}
{\thanks {\vspace{-0.5cm}\hrule
				\vspace{0.1cm} This work was made possible, in part, by grant NPRP \#NPRP12S-0313-190348 from the Qatar National Research Fund (a member of The Qatar Foundation). The statements made herein are solely the responsibility of the authors.
		}}
}

\begin{document}
\maketitle
\thispagestyle{empty}
\begin{abstract}
\boldmath Crowd management technologies that leverage computer vision are widespread in contemporary times. There exists many security-related applications of these methods, including, but not limited to: following the flow of an array of people and monitoring large gatherings. In this paper, we propose an accurate monitoring system composed of two concatenated convolutional deep learning architectures.
The first part called Front-end, is responsible for converting bi-dimensional signals and delivering high-level features. The second part, called the Back-end, is a dilated Convolutional Neural Network (CNN) used to replace pooling layers. It is responsible for enlarging the receptive field of the whole network and converting the descriptors provided by the first network to a saliency map that will be utilized to estimate the number of people in highly congested images. We also propose to utilize a genetic algorithm in order to find an optimized dilation rate configuration in the back-end. The proposed model is shown to converge 30\% faster than state-of-the-art approaches. It is also shown that it achieves 20\% lower Mean Absolute Error (MAE) when applied to the Shanghai data~set.
\end{abstract}
\section{Introduction}
The study of the collective behavior of mass gatherings has attracted much research interest and governmental investment in order to efficiently monitor a crowd to reduce the risk of crowd-related disasters \cite{dollar2011pedestrian}. With the improvement of video surveillance infrastructure, video-based human-behavior motion analysis has become the new state-of-the-art.
Various techniques have been developed to deal with highly congested images and accurately estimate the number of objects in these scenes. These heuristics vary from directly counting the number of people present in the target images to providing density maps that reflect the distribution of the targets in given images. The reason behind choosing a density map as an output is that dealing with these feature maps is more accurate as they provide more insights on how the models perceive different clusters in the frame.

The majority of the work in congested scenes can be divided into four groups:
The first is a \textit{detection-based approach}. It relies on a moving window over the image that uses a trained algorithm to extract low level features from the object in question, e.g. the case of human detection using Haar features and Histogram Oriented Gradient (HOG). However, these types of straightforward models perform poorly when it comes to highly dense images where the targets are compact and most of times are partly visible.
The second method is a \textit{regression-based approach}. In order to remedy the errors of the previous method while dealing with dense scenes, researchers developed regression-based techniques to utilize the relations among descriptive features extracted from different regions and then aggregate them to estimate the total count in these images. Low level features can be fed to regression models such as Scale Environment Feature Transform (SIFT) and use interest points to estimate the count \cite{idrees2013multi}.
The third technique is a \textit{density-based method}. As a matter of fact, one critical aspect of images that is left unattended by regression-based methods is called saliency.
It describes the conspicuity at every position in the visual field by a finite number and guides the selection of important patches, based on the spatial distribution of saliency. To remedy this weakness, this method proposes a linear mapping between local features and the density map \cite{lempitsky2010learning} and a nonlinear mapping using a random forest regressor to vote for densities of multiple target objects\cite{pham2015count}. This mapping exists between the patch features and the relative locations of all objects inside the patch. 

The final method employs \textit{Convolutional Neural Networks} (CNN);
this CNN based methods have been widely used to deal with congested scenes  \cite{ simonyan2014very}, due to their ability to learn the semantic information of the image. Currently, there are two major strategies to deploy CNNs in such problem-sets. The first is to directly learn mapping between images and the count by directly training the network on the images and their respective counts. 
\begin{figure}
  \includegraphics[width=\linewidth,height=8 cm,keepaspectratio]{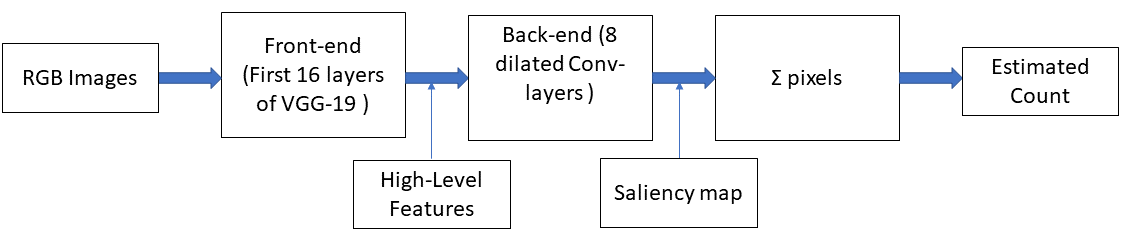}\vspace{-0.2cm}
  \caption{The components of our proposed architecture and the types of output at each phase.}\vspace{-0.6cm}
  \label{fig:plot0}
\end{figure}
\space
The second aims at predicting density maps and then sum all the pixels to obtain the count. These models are trained to optimize their density maps instead of directly optimizing the predicted number of people since the former preserves more information than the latter. One of the previously used architectures are called Multi-column CNNs (MCNNs) \cite{zhang2016single}, which stack multiple columns of CNNs and utilize different kernels in order to extract deep descriptors from the frame.
Another architecture used in \cite{boominathan2016crowdnet} is called Crowd-Net. It uses two parallel networks: one is a shallow network and the other is deep and combines both outputs to construct a feature map and sum its pixels to generate the total count.

The main focus in this paper is to provide a more accurate solution by using high-level features and deep neural networks without losing too much information. We propose an architecture that can be decomposed into two concatenated convolutional neural networks as shown in Fig.~\ref{fig:plot0}, the first part is called \textbf{Front-end} and the second part is named \textbf{Back-end}. 
The former is responsible for extracting high-level features from the input images while the latter is used to concatenate and aggregate the deep descriptors feature extractor in order to provide deep information about the saliency to estimate the number of people in these highly dense images.

\begin{figure}
  \includegraphics[width=\linewidth]{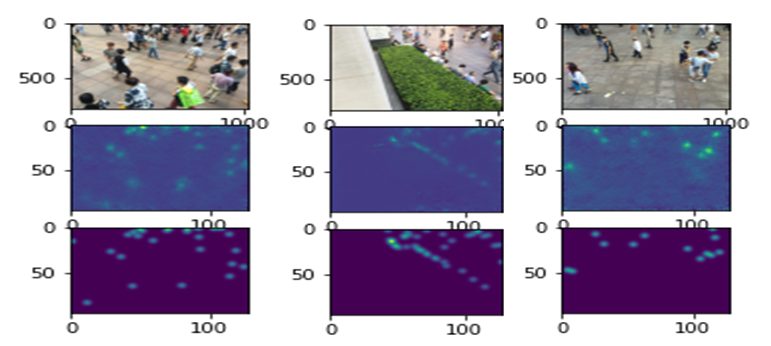}\vspace{-0.2cm}
  \caption{Example: 
  Pictures in the first row are the real images, pictures in the second row are the predicted density maps, and the pictures in the third row are the true density maps.}\vspace{-0.5cm}
  \label{fig:plot6}
\end{figure}

In this paper, we propose an end-to-end architecture that not only estimates accurately the number condensed objects/persons in given frames, but also provides a precise density map that shows the location of the clusters and that can be used for crowd behaviour studies as demonstrated in Fig.~\ref{fig:plot6}. Our proposed model not only outperforms the state-of-the-art in terms of speed of convergence by taking less than 30\% number of epochs to reach stagnation, it also scores an MAE lower by 20\% than the state-of-the-art's on Shanghai dataset and that is due our method of optimizing the choice of the hyper-parameters by using a genetic algorithm as our search technique.
\vspace{-0.2cm}
\section{Methodology and Architecture}
\label{Sec2}
Our proposed solution relies on density estimation instead of directly predicting the count, therefore our first step is to generate each image's true density map.
\subsection{Ground Truth Generation}
We first have to generate the ground truth values from the original images. A feature map is generated with 1/8 the size of the initial image while considering the spatial distribution of the pixels in the image. This can be achieved by blurring each head using a normalized Gaussian kernel (normalized to 1). The geometry-adaptive filter is denoted by $F(t)$ and expressed as follows:
\begin{equation}
    F(t)={\sum_{n=1}^{N}\delta(t-t_{n})\cdot G_{\sigma _{n} }}, \text{ where} \indent  {\sigma_{n} }=\beta_{n}\cdot\bar{d_{n}.}
\end{equation}
Given $N$ targets in the image, for each targeted object $t_{n}$ in the ground truth $\delta$, we denote by $\bar{d_{n}}$ the average distance of $k$ nearest neighbors in each of the $N$ target images. To generate the density map, we convolve $\delta(t-t_{n})$ with a Gaussian kernel $G_{\sigma _{n}}$, with a standard variation $\sigma_{n}$, where $t$ is the position of pixel in each image. In our experiment, we follow the configuration in \cite{zhang2016single} where $\beta_{n}$ = 0.3 and $k$ =3.
After generating the true feature maps for each image in our dataset, the next step is finding the best feature extractor we can use from the numerous architectures whilst fixing the same back layers and with the same hyper-parameters' configuration.
\begin{figure}[t]
  \includegraphics[width=\linewidth]{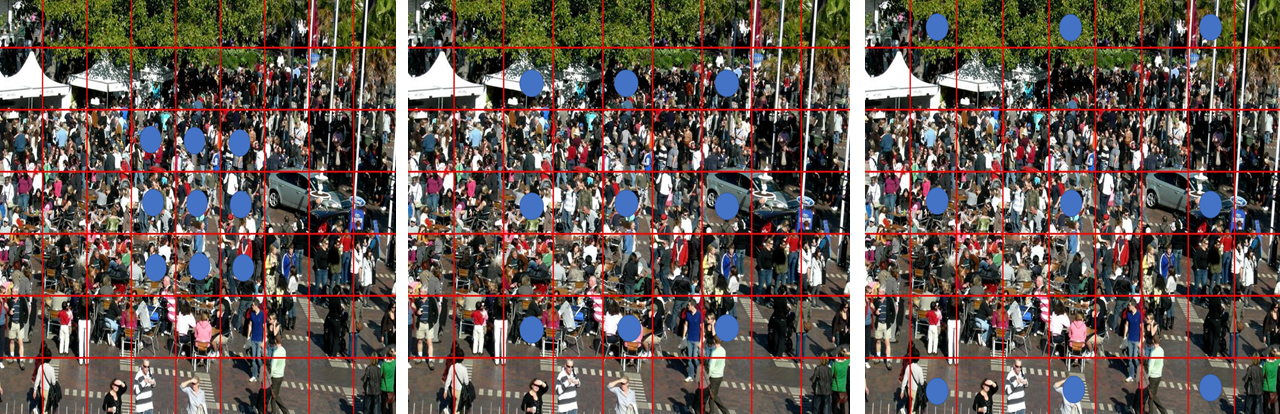}\vspace{-0.2cm}
  \caption{Resulting convolutions with a kernel the size $3\times3$ after using a dilation rate of 1, 2, and 3, in the far left, in the middle, and in the far right, respectively.}
  \label{fig:plot4}
\end{figure}
\subsection{Front-End Development}
In this case, we adopt the Visual Group Geometry 19 (VGG-19) pretrained on ImagNet~\cite{deng2009imagenet}, We select this model due to its light architecture, flexibility, and ease of use for transfer learning. We use multiple layers of this architecture except the output layers in order to preserve pretrained model weights.. Although this model contains 19 layers, we select only the first 12 layers since it gave the best results based on a complexity/perfomance tradeoff.\vspace{-0.2cm}
\subsection{Back-End Development}
We propose to make few adjustments to the Back-end in order to make it compatible with the pre-trained VGG-19 architecture. We add \textbf{two} more layers on top of the Back-end used in \cite{li2018csrnet} with \textbf{1024} \textbf{channels} in order to better use the highly granular input coming from the Front-end.
\subsubsection{Dilated Convolutions}
Although pooling layers are widely used for maintaining invariance and controlling over-fitting, they also dramatically decrease the spatial resolution. In other words, pooling layers lead to losses of spatial information.
A spatial dilated convolution can be defined as follows: 
\begin{equation}
    z(m,n)=\sum_{i=1}^{M}\sum_{j=1}^{N}{x(m+r\cdot{i},n+r\cdot{j})\cdot{w(i,j)}},
\end{equation}
where $z(m, n)$ is the output of a dilated convolution from input x(m, n) and a kernel $w(i, j)$ with the length and the width of $M$ and $N$ respectively. The parameter $r$ is the dilation rate. If $r=1$, a dilated convolution turns into a normal convolution and if the value of $r$ is greater than 1 then, our image will be convoluted to a kernel with a stride equal to $r-1$ as shown in~Fig.~\ref{fig:plot4}.
\subsubsection{Genetic Algorithm for Dilation Rates Optimization}

The objective is to find the best set of dilation rates for the new Back-end. Therefore instead of using the classic Grid Search method, we propose the use of a genetic algorithm search, due to its speed and efficiency when it comes to finding the best candidates in a given environment.
Genetic algorithms are a search technique that simulate the process of natural selection, fittest candidates reproduce and pass their offspring to the next generation, replacing the unfit candidates.
The process starts with an initial set of candidate solutions, called the  \textbf{Population}. 
Each candidate solution (member) is comprised of a vector encoding refered to as chromosomes. Each chromosome is formed by a number of genes. In our case, a gene is a dilation rate in each convolution layer and a chromosome is the set of dilation rates for each Back-end in each network. This algorithm is characterized by repeating the following five instructions: (a) population generation, (b) fitting each member of the population, (c) selection of fittest individuals, (d) cross-over and breeding of the survivals, and (e)  randomly mutating newly bred networks. The genetic algorithm for dilation rate optimization is detailed in Algorithm~\ref{alg:pseudoPSO}.
\setlength{\textfloatsep}{0.1cm}
\setlength{\floatsep}{0.1cm}
\begin{algorithm}[t]
\caption{Genetic Alg. for Dilation Rate Optimization}
\label{alg:pseudoPSO}
\begin{algorithmic}[1]
\State Choose the number of generations and the number of populations in each generation;  
\State Choose mutation\_rate, retain\_rate, learning\_rate;
\State Define $L$ as the list of possible dilation rates;
\State Networks$\gets$\{\};
\For{$i_{gen}$ in range(1,Generation)} 
\If{Networks $== \{\}$}
    \For{$j$ in range(1,Population)}
        \State Param\_dict$\gets${\textbf{Random\_choice}($L$, \text{for $i$ in range(1,8)})};
        \State Networks($j$)$\gets${\textbf{Create\_network}(Param\_dict, learning\_rate, Batch-size)};
    \EndFor
\EndIf    
\For{$i_{pop}$ in range(1,Population)}

    \State \textbf{Train\_Network}(Networks($i_{pop}$));
    \State MAE($i_{pop}$)$\gets$\textbf{Evaluate\_MAE}(Networks$(i_{pop})$);
\EndFor
\State Parents=Select best retain\_rate performing networks in Networks;\Comment{minimizing MAE}
\State Children$\gets$\textbf{CrossOver\_Breed}(\text{Parents' Param\_dict});
\If{rand() $<$  Mutation Rate} 
   \State Children$\gets$\textbf{Randomly\_Mutate}(Children's Param\_dict);
\EndIf
\State Networks$\gets$\{Parents,Children\};
\EndFor
\State Select the best performing network in Networks in the final generation;\Comment{Minimizing MAE}
\end{algorithmic}
\end{algorithm}
\section{Experimental Approach and Results}
In this section, we present the experimental approach, evaluate the proposed deep learning architecture, and compare its performance with the state-of-the-art.

\subsection{Shanghai Dataset and Data Augmentation}
The Shanghai Dataset is comprised of nearly 1,200 images with 330,000 labeled heads. It is the largest crowd counting dataset in terms of the number annotated heads. This dataset consists of two parts: Part A and B. Images in Part A are randomly collected from the Internet; most of them have a large number of people. Part B are captured from crowded streets of metropolitan areas in Shanghai. We crop nine patches from each frame at different positions with 1/4 size of the original image. The first four patches contain four quarters of the image without overlapping while the other five patches are randomly cropped from the input image. Afterwards, we mirror the patches so that we double the training set.
\subsection{Evaluation Metrics}
In this paper, we utilize both Mean Absolute Error (MAE) and Mean Squared Error (MSE) as performance metrics to evaluate our proposed approach. The MAE and MSE are defined as follows: 
\begin{align}
 &\text{MAE}=\frac{1}{S}{\sum_{k=1}^{S}\vert{C_{k}-C_{k}^{GT}}\vert}, \text{MSE}=\sqrt{\frac{1}{S}\sum_{k=1}^{S}\vert{C_{k}-C_{k}^{GT}}\vert^2},\notag\\
&\text{where } C_{k}=\sum_{i=1}^{W}\sum_{j=1}^{H}{Z_{i,j}},
\end{align}
and $W$ and $H$ are the width and height of a predicted feature map. The parameter $Z_{i,j}$ denotes the value of the pixel at the position $(i,j)$. 

\begin{figure}
  \includegraphics[width=\linewidth,height=40pt]{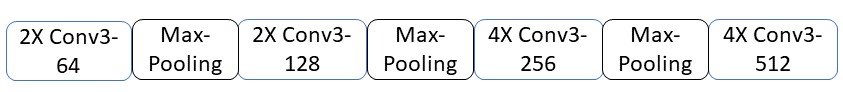}\vspace{-0.2cm}
  \caption{Layers loaded from pre-trained VGG-19 on ImageNet.}\vspace{-0.0cm}
  \label{fig:plot2}
\end{figure}
\subsection{Training Pipeline}
In all the experiments, we use transfer learning i.e., we initialize the Front-end with pre-trained layers from VGG-19, truncate the first twelve layers and link them with a fixed Back-end with the same architecture and hyper-parameter configuration. Next we initialize the backend weights with gaussian noise that has a standard deviation of 0.01, and utilize a Stochastic Gradient Descent (SGD) optimizer with a learning rate of $ = 10^{-7}$, 400 epochs and the Euclidean distance as the objective function, which is given as follows:
\begin{equation}
    \text{L(T)}=\frac{1}{2D}\sum_{i=1}^{D}{\vert\vert{Y(X_{i};T) - Y^{GT}_{i}}\vert\vert^2_{2}}.
    \label{eq:4}
\end{equation}
In \eqref{eq:4}, $D$ is the size of training batch and $Y(X_{i}; T)$ is the output generated by models with parameters $T$. The notation $X_{i}$ represents the input image while $Y^{GT}_{i}$ is the ground truth result of the input image $X_{i}$. The notation $\|.\|_{2}$ denotes the L2 norm. 
For the genetic algorithm search, we utilize a population size of 7 and we evolve the population over 7 generations. The survival rate is set to 0.4 (i.e. the fittest 40\% of the population reproduces to round out the population in each successive generation), and we set the mutation rate to 0.2 (i.e. each member of the population can mutate up to 20\% of its genes).
As mentioned earlier, the aim is to improve the performance of the whole model by finding the best feature extractor and the best Back-end for this use-case. Due to the hardware limitation, we choose the Shanghai dataset as the reference data and we choose VGG-19 pre-trained on Image-net dataset shown in Fig.~\ref{fig:plot2} as the feature extractor. 
Despite the connection of the VGG-19 layers with the same back-end as CSRNet in~\cite{li2018csrnet}, it outperforms the front-end in the state-of-the art.  Moreover, when used with the proper Back-end, it gives better results on the Shanghai dataset. 
\begin{figure}[t]
  \includegraphics[height=4cm,width=8cm]{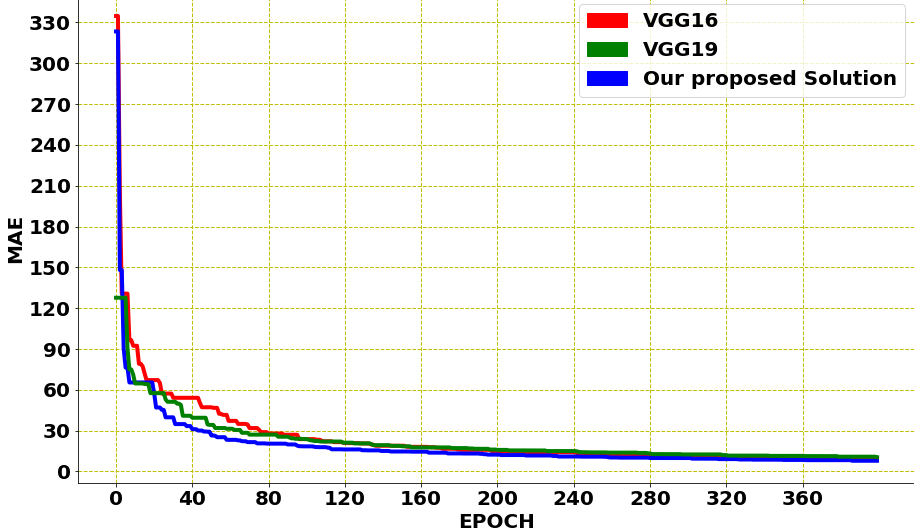}\vspace{-0.2cm}
  \caption{Validation MAE from training on the Shanghai dataset.} 
  \label{fig:plot3}
\end{figure}

We train three networks. The first is the afromentionned CSRNet, the second is a network constructed by using the same layers used in CSRNet's \textbf{front-end} but from VGG-19 instead of VGG-16 with the same back-end as the first network. The third is a combination between the front-end from the second architecture concatenated by our own back-end. The results from the second network show that replacing pretrained layers from VGG-16 with those in VGG-19 results in enhancing the model, thus we chose to use transfer learning with VGG-19. We kept on adding layers from that pretrained network untill we reached the configuration shown in Fig.~\ref{fig:plot0}.
\begin{table}
\begin{center}
\caption{Validation MAE each 100 epoch}
\label{tab1}
 \begin{tabular}{||c c c c||}
 \hline
 Epoch & VGG-16 & VGG-19 & Our model \\ [0.5ex] 
 \hline\hline
 100 & 57.290 & 44.988 & 42.187 \\ 
 \hline
 200 & 38.084 & 32.038 & 28.553 \\
 \hline
 300 & 29.763 & 26.114 & 21.652 \\
 \hline
 400 & 24.924 & 24.470 & 19.144 \\
\hline

\end{tabular}
\vspace{-0.2cm}
\end{center}
\end{table}
This model outperforms CSRNet in terms of both MAE and in convergence speed, as less epochs are required to reach stagnation. Fig.~\ref{fig:plot3} illustrates the mean value of MAE on the validation data after each epoch; it shows that our network reached its stagnation around 270 epochs, which is 30\% faster than that of the state-of-the-art which was around 390 epochs. Our proposed solution converges at around 200 epochs. 

\begin{figure}
	\centering
	\begin{subfigure}{0.5\textwidth}
 	\centerline{\includegraphics[width=8cm,height=3.5cm]{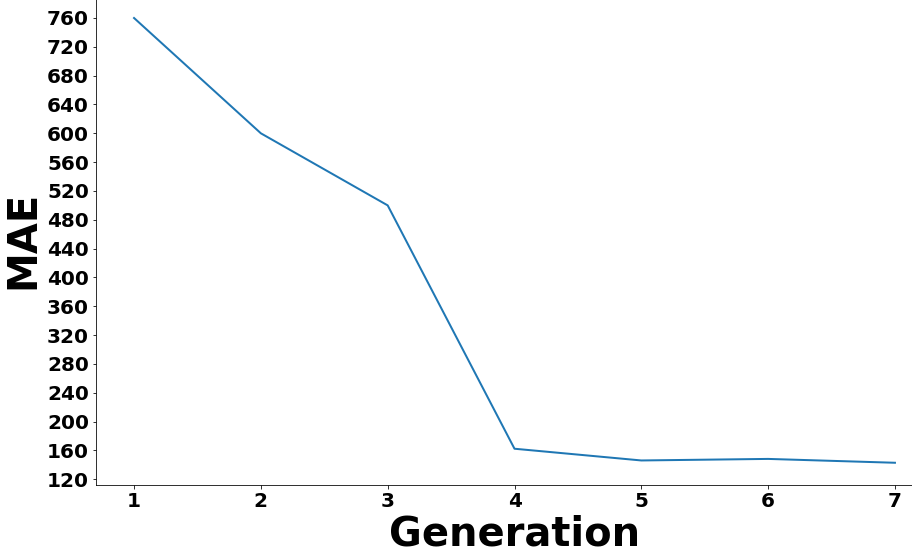}}
		\caption{MAE of the best performing networks in each generation (trained on 20 epochs).}
		\label{fig:plot9}
	\end{subfigure}
\\	\begin{subfigure}{0.5\textwidth}
	\centerline{\includegraphics[width=8cm,height=3.5cm]{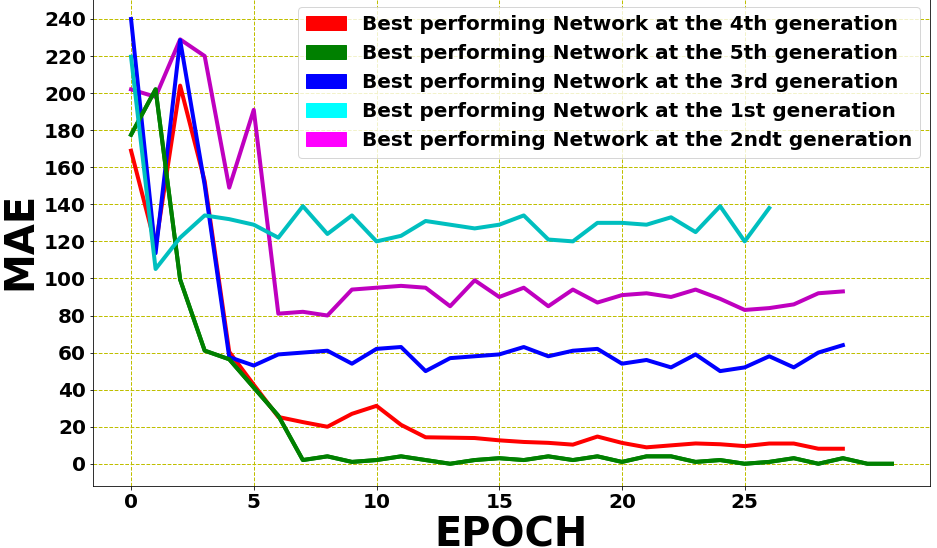}}
		\caption{MAE rolling mean with a window size of 10 epochs of the best networks trained on 300 epochs.}
		\label{fig:plot10}
	\end{subfigure}\vspace{-0.2cm}
	\caption{Genetic search output metrics.}
\end{figure}

Moreover, in Table~\ref{tab1}, we summarize the performance of each network after 100 iterations by measuring the MAE of training at the mentioned epochs.
The mean of the validation MAE during the last 40 epochs of our proposed network was equal to 6.458, and the MAE of the state-of-the-art was equal to 7.7496 which is 20\% higher than our MAE.
Our aim is to find the best set of dilation rates by using the Genetic Algorithm. We choose to limit the space of these parameters from 2 to 5, we elevate the batch size from 1 to 8. We use bi-linear interpolation in order to stack the tensors and their respective targets in the same batch, and we reduce the number of epochs in each training routine from 400 to 20. All of these measures linearly affected the performance all the models, however, it is necessary to use them since we intended to train 7 candidates per generation as shown by Fig.~\ref{fig:plot10}, where we trained the best networks in each generation on 300 epochs and with batch-size equal to one in order to compare the networks' performance in optimal conditions.

From Fig. \ref{fig:plot9}, we see that in the third generation, the algorithm converges to a new candidate, and in the fourth generation it converges to the network with the best performance. This is either because the mutations in later generations are less effective, or that the network constructed in the fourth generation is the best possible network based on the initial conditions.
\vspace{-0.3cm}
\section{Conclusion}
In this paper, we proposed an-easier-to-train deep learning architecture that achieves 20\% lower MAE and 30 \% convergence speed gain compared to state-of-the-art techniques for mass gathering detection. We proved as well that the use of the genetic algorithm was an efficient method to tune the hyper-parameter configuration via evolutionary search methods in order to find more effective networks. Moving forward, we will consider the applications of genetic algorithms to tune the learning rate, along with the remaining hyper-parameters.
\vspace{-0.3cm}

\bibliographystyle{ieeetr}
\bibliography{References}

\end{document}